\documentclass{article}

\usepackage{iclr2027_conference,times}
\iclrfinalcopy

\usepackage[utf8]{inputenc} 
\usepackage[T1]{fontenc}    
\usepackage{hyperref}       
\usepackage{url}            
\usepackage{booktabs}       
\usepackage{amsfonts}       
\usepackage{amsmath}        
\usepackage{amssymb}        
\usepackage{mathtools}      
\usepackage{nicefrac}       
\usepackage{microtype}      
\usepackage{xcolor}         
\usepackage{graphicx}       
\usepackage{enumitem}       
\usepackage{algorithm}      
\usepackage{algorithmic}    
\usepackage{multirow}
\usepackage{wrapfig}
\usepackage[most]{tcolorbox}

\title{\textsc{SkillGraph}: Skill-Augmented Reinforcement Learning for Agents via Evolving Skill Graphs}

\author{%
  Xiaoyuan Li$^{1}$ \quad
  Moxin Li$^{3}$ \quad
  Keqin Bao$^{2}$  \quad
  Yubo Ma$^{2}$ \quad \\
  \textbf{Wenjie Wang}$^{1}$ \quad 
  \textbf{Fuli Feng}$^{1}$ \quad 
  \textbf{Dayiheng Liu}$^{2}$ \quad \\
  $^{1}$University of Science and Technology of China \quad
  $^{2}$Alibaba Group \quad \\
  $^{3}$National University of Singapore \\
}

\newcommand{\answerTODO}[1][]{\textcolor{red}{\bf [TODO]}}
\newcommand{\justificationTODO}[1][]{\textcolor{red}{\bf [TODO]}}

\begin{document}

\maketitle
\begin{abstract}
Skill libraries enable large language model agents to reuse experience from past interactions, but most existing libraries store skills as isolated entries and retrieve them only by semantic similarity. 
This leads to two key challenges for compositional tasks. 
Firstly, an agent must identify not only relevant skills but also how they depend on and build upon each other. Secondly, it also makes library maintenance difficult, since the system lacks structural cues for deciding when skills should be merged, split, or removed.
We propose \textsc{SKILLGRAPH}, a framework that represents reusable skills as nodes in a directed graph, with typed edges encoding prerequisite, enhancement, and co-occurrence relations. Given a new task, \textsc{SKILLGRAPH} retrieves not just individual skills, but an ordered skill subgraph that can guide multi-step decision making. The graph is continuously updated from agent trajectories and reinforcement learning feedback, allowing both the skill library and the agent policy to improve together. Experiments on ALFWorld, WebShop, and seven search-augmented QA tasks show that SKILLGRAPH achieves state-of-the-art performance against memory-augmented RL methods, with especially large gains on complex tasks that require composing multiple skills.
\end{abstract}

\section{Introduction}
\label{sec:introduction}

Large Language Model (LLM) agents have shown strong capabilities in complex interactive tasks, including web navigation~\citep{yao2022webshop}, embodied household manipulation~\citep{shridhar2020alfworld}, and tool-augmented question answering~\citep{yao2022react}. 
Yet most agents treat task as episode~\citep{yao2022react, shinn2023reflexion}, struggling to learn from past successes or failures even when structurally similar problems have been encountered~\citep{xia2025skillrl}.
Since many tasks share recurring subproblems and compositional action patterns, an agent that can \emph{learn from experience}---extracting reusable knowledge from past interactions---would avoid redundant exploration, transfer strategies to similar tasks, and progressively build up the ability to solve more complex problems.

To reuse experience, a common approach is to maintain a \emph{skill library}, which stores reusable units of knowledge for solving recurring subproblems~\citep{wang2023voyager, zhao2024expel, xia2025skillrl}. 
A skill can be either manually designed by humans~\citep{xu2026agent} or automatically acquired from agent experience---for instance, by distilling successful trajectories into natural language~\citep{zhao2024expel, xia2025skillrl} or executable programs~\citep{wang2023voyager}. 
Compared with manually crafted skills, automatically acquired skills are more scalable and can continuously expand as the agent encounters new tasks and environments. 
Therefore, we focus on automatically acquiring skills from interaction trajectories.

Despite their promise, existing skill libraries are often organized as flat collections, where each skill is stored as an independent entry and retrieved mainly by semantic similarity~\citep{xia2025skillrl, zhao2024expel, liu2025simplemem}. 
This ignores the fact that skills are inherently related: some skills are prerequisites for others, some enhance others, and some frequently co-occur in successful trajectories. 
As a result, flat libraries suffer from two key limitations. 
First, \textbf{retrieval is not compositional}. 
Complex tasks often require an ordered sequence of skills; for example, a ``heat and place'' task in ALFWorld may require locating an object, picking it up, heating it with an appliance, and then placing it at the target destination. 
A flat Top-$K$ retriever can return relevant skills, but it does not indicate their dependencies or execution order. 
Second, \textbf{skill updates are not structured}. 
When skills are maintained independently, the library lacks explicit evidence for merging redundant skills, splitting overly broad skills, deprecating obsolete skills, or strengthening useful relations between skills~\citep{xu2026agent}.
These limitations suggest that the core problem is not only how to acquire skills, but also how to  \emph{organize, retrieve, and update} them. 
If inter-skill relations are explicitly represented, retrieval can produce dependency-aware skill sequences rather than unordered hints, and both individual skills and their relations can be updated in a principled way.

Motivated by this, we propose \textsc{SkillGraph}, a framework that organizes skills into a structured graph and co-evolves it with the agent's policy through reinforcement learning (RL). 
In \textsc{SkillGraph}, nodes represent skills distilled from trajectories, while typed edges capture relations such as prerequisite, enhancement, and co-occurrence. \textsc{SkillGraph} consists of three stages. 
First, \textbf{graph construction} builds an initial skill graph from interaction trajectories, making inter-skill relations explicit. 
Second, \textbf{graph-aware retrieval} starts from task-relevant seed skills, expands along graph edges, and orders retrieved skills according to their dependencies, producing a coherent skill sequence for decision-making. 
Third, \textbf{graph evolution} updates the graph during training by refining skill nodes and adjusting edge relations according to skill usage and success rate. 
Together, these stages form a closed loop: the skill graph provides structured guidance for policy learning, while the improving policy generates new trajectories that further refine the graph.

\begin{figure}[t]
    \centering
    \includegraphics[width=\linewidth]{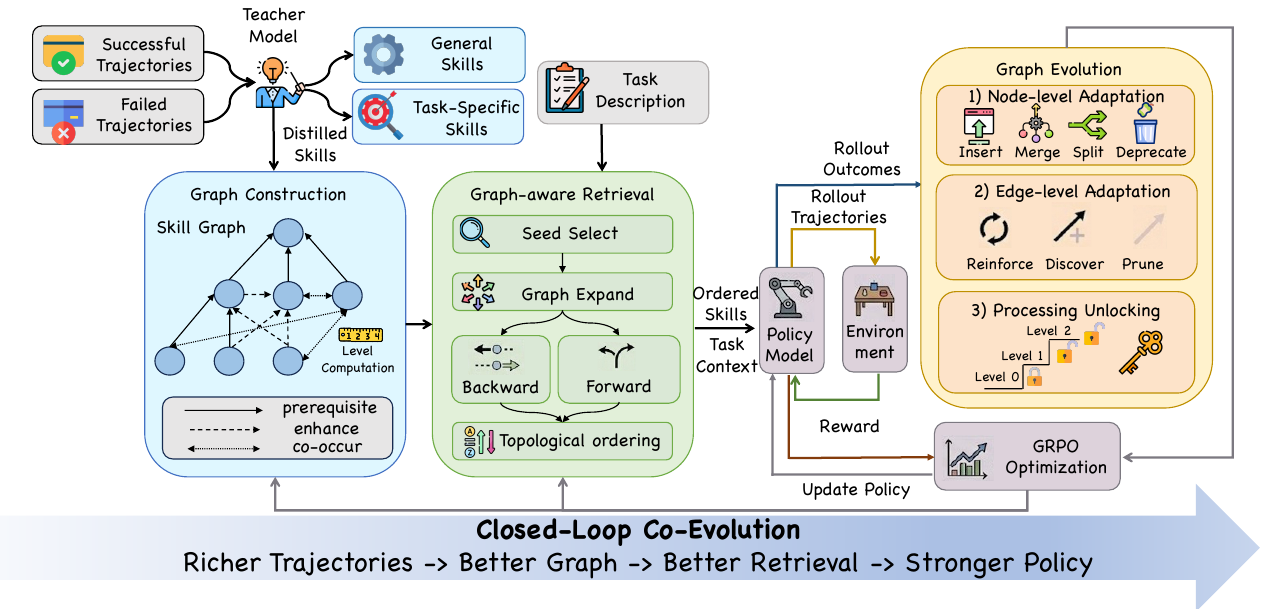}
    \vspace{-20pt}
    \caption{Overview of \textsc{SkillGraph}. The skill graph and the agent's policy \emph{co-evolve} through a closed loop: \textbf{(1)}~graph construction distills skills and their typed relations (prerequisite, enhancement, co-occurrence) from trajectories; \textbf{(2)}~graph-aware retrieval traverses these relations to produce dependency-ordered skill sequences that guide the policy; \textbf{(3)}~graph evolution uses training feedback to refine skill nodes, adjust edge weights, and restructure the graph, which in turn improves future retrieval and policy learning.}
    \label{fig:overview}
    \vspace{-20pt}
\end{figure}

Empirically, we evaluate \textsc{SkillGraph} on ALFWorld~\citep{shridhar2020alfworld}, WebShop~\citep{yao2022webshop}, and seven search-augmented question answering tasks~\citep{kwiatkowski2019nq}, covering embodied manipulation, web navigation, and information retrieval. 
Experimental results show that \textsc{SkillGraph} achieves state-of-the-art performance across all benchmarks, with especially strong gains on complex multi-step tasks requiring skill composition. 
Further analysis shows that the graph structure improves skill reuse, reduces redundancy compared with flat libraries, and enables transfer of compositional knowledge from simpler tasks to more complex ones.

Our main contributions are summarized as follows:
\begin{itemize}[leftmargin=*, nosep]
    \item We propose a graph-structured formulation of skill library for LLM agents, where skills are connected by explicit prerequisite, enhancement, and co-occurrence relations.
    \item We introduce \textsc{SkillGraph}, a closed-loop framework that supports dependency-aware skill retrieval and structured skill updates during RL.
    \item We conduct experiments on ALFWorld, WebShop, and seven search-augmented QA tasks, demonstrating state-of-the-art performance and substantial gains on complex multi-step tasks.
\end{itemize}
\section{Related Work}
\label{sec:related}

\paragraph{Memory mechanisms in agents.}
External memory helps LLM agents reuse experience beyond the context window. Early methods store raw trajectories as examples~\citep{zhao2024expel, chhikara2025mem0}, while later work compresses experience into summaries or knowledge entries~\citep{fang2025memp, liu2025simplemem, ouyang2025reasoningbank, tang2025agentkb}. Recent studies further apply RL directly to agent knowledge structures: MemRL~\citep{zhang2026memrl} performs runtime RL on episodic memory, MemEvolve~\citep{zhang2025memevolve} meta-evolves memory systems, Mem-$\alpha$~\citep{wang2025mem_alpha} learns memory construction policies, and EvolveR~\citep{wu2025evolver} co-adapts the policy and memory bank. In contrast, \textsc{SkillGraph} represents experience as explicit skill abstractions with typed dependencies and evolves this structure jointly with the policy.

\paragraph{Graph structures for LLMs.}
Graph structures have been used to enhance LLM reasoning~\citep{besta2024got}, retrieval~\citep{edge2024graphrag}, and agent planning~\citep{huang2024graphplan}. For agent skills specifically, Graph of Skills~\citep{liu2026gos} proposes dependency-aware structural retrieval. \textsc{SkillGraph} differs in that it jointly evolves the graph topology and the policy through RL, enabling the graph to adapt continuously via training feedback rather than remaining static after construction.

\paragraph{Agent skill evolution.}
Agentic skills can compact reusable strategies for subtasks. Voyager~\citep{wang2023voyager} accumulates executable code skills, and ExpeL~\citep{zhao2024expel} distills transferable strategic experience from trajectories. Most closely related, SkillRL~\citep{xia2025skillrl} co-evolves a hierarchical skill bank with the agent's policy through recursive RL. \textsc{SkillGraph} builds on this line by elevating the flat skill bank into a structured dependency graph, enabling typed relational modeling and topology evolution throughout training.

\section{\textsc{SkillGraph}}
\label{sec:method}

We present \textsc{SkillGraph}, a framework that organizes agent skills as a directed dependency graph and co-evolves the graph with the agent's policy through RL. The key insight is that explicitly modeling inter-skill relations enables two mutually reinforcing capabilities: \emph{structured retrieval} that produces dependency-aware skill sequences for compositional planning, and \emph{principled evolution} that uses training feedback to refine both individual skills and their relations. As illustrated in Figure~\ref{fig:overview}, the framework consists of three stages---graph construction (Section~\ref{sec:construction}), graph-aware retrieval (Section~\ref{sec:retrieval}), and graph evolution (Section~\ref{sec:evolution})---integrated into a closed-loop training procedure (Section~\ref{sec:training}).

\subsection{Graph Construction}
\label{sec:construction}

The first step is to build a skill graph that makes inter-skill relations explicit, providing the structural foundation for both retrieval and evolution.

\paragraph{Skill distillation.}
We collect trajectories by rolling out the base policy $\pi_{\text{base}}$ in the environment. A teacher language model $\mathcal{M}$ then distills successful trajectories $\tau^+$ and failed trajectories $\tau^-$ into two types of skills: \emph{general skills}, which capture domain-independent reasoning strategies applicable across tasks (e.g., ``verify each sub-goal before proceeding''), and \emph{task-specific skills}, which encode strategies tied to particular task types (e.g., ``check the microwave for heated objects''). Each skill is represented as a compact record containing a title, a core principle describing the strategy, an applicability condition, and a category label indicating its type.

\paragraph{Graph structure.}
The distilled skills form the node set $\mathcal{V}$ of a directed graph $\mathcal{G} = (\mathcal{V}, \mathcal{E})$, where $\mathcal{E}$ denotes the edge set. To capture how skills relate to one another, we define three typed edges:
\begin{itemize}[leftmargin=*, nosep]
    \item \textbf{Prerequisite} ($A \xrightarrow{\texttt{prereq}} B$): skill $A$ must be applied before skill $B$.
    \item \textbf{Enhances} ($A \xrightarrow{\texttt{enhance}} B$): general skill $A$ improves the effectiveness of task-specific skill $B$.
    \item \textbf{Co-occurs} ($A \xleftrightarrow{\texttt{co\_occur}} B$): skills $A$ and $B$ frequently appear together in successful episodes.
\end{itemize}
Each edge $e \in \mathcal{E}$ carries a weight $w(e) \in [0, 1]$ reflecting the strength of the relation, which is dynamically adjusted during training. Each node $v \in \mathcal{V}$ maintains running statistics---usage count $n_\text{use}(v)$, success count $n_\text{succ}(v)$, and empirical success rate $\hat{p}(v) = n_\text{succ}(v) / n_\text{use}(v)$---that drive both evolution decisions and progressive unlocking in Section~\ref{sec:evolution}. Based on the directed prerequisite and enhancement edges, each node is assigned a topological level $\ell(v)$ indicating its position in the dependency hierarchy: level-0 skills have no prerequisites, while higher-level skills depend on lower-level ones. Details of edge initialization and level computation are provided in Appendix~\ref{app:entire_pipeline}.

\subsection{Graph-Aware Retrieval}
\label{sec:retrieval}

Flat skill retrieval returns a set of individually relevant skills but ignores their dependencies, making it inadequate for tasks that require ordered skill composition. To address this, we design a graph-aware retrieval procedure that traverses the skill graph to produce a dependency-respecting sequence of skills. Given a task description $d$ with task type $t(d)$, retrieval proceeds in three steps.

\paragraph{Seed selection.}
We first identify task-relevant entry points from the currently active skill set $\mathcal{V}_{\text{active}} \subseteq \mathcal{V}$, which contains skills that have been progressively unlocked (see Section~\ref{sec:unlocking}). From $\mathcal{V}_{\text{active}}$, we select all general skills and task-type-matched skills as seed nodes, where $\mathcal{R}$ denotes a retrieved subset of skill nodes:
\begin{equation}
    \mathcal{R}_{\text{seed}} = \left\{ v \in \mathcal{V}_{\text{active}} : \mathrm{category}(v)=\texttt{general} \;\vee\; \mathrm{category}(v)=t(d) \right\}.
    \label{eq:seed_selection}
\end{equation}

\paragraph{Graph expansion.}
Starting from the seed set $\mathcal{R}_{\text{seed}}$, we expand in two complementary directions to recover the full dependency context:
\begin{itemize}[leftmargin=*, nosep]
    \item \emph{Backward expansion} traverses incoming prerequisite edges via breadth-first search (BFS) up to a maximum depth $D$, producing the backward-expanded set $\mathcal{R}_{\text{BFS}}$ that recovers foundational skills the seeds depend on but that may belong to other task categories.
    \item \emph{Forward expansion} explores outgoing edges via beam search with beam width $B$, producing the forward-expanded set $\mathcal{R}_{\text{beam}}$. Each candidate node $v$ receives an expansion score $\sigma(v)$ propagated from its predecessors: $\sigma(v) = \max_{u \in \text{parents}(v)} \sigma(u) \cdot w(u,v)$, where seed nodes are initialized with $\sigma = 1$. This prioritizes skills connected by well-validated relations.
\end{itemize}

\paragraph{Topological ordering.}
The union of seeds, backward-expanded, and forward-expanded skills is topologically sorted according to the graph's dependency edges, producing an ordered skill sequence:
\begin{equation}
    \mathcal{R}_{\text{ret}} = \text{TopoSort}_{\mathcal{G}}\!\left(\mathcal{R}_{\text{seed}} \cup \mathcal{R}_{\text{BFS}} \cup \mathcal{R}_{\text{beam}}\right).
    \label{eq:topo_sort}
\end{equation}
This sequence, capped at $K_{\max}$ skills, is prepended to the task prompt as structured guidance for the policy. Because the ordering reflects dependency relations, the agent receives skills in a natural simple-to-complex order that mirrors how sub-tasks should be composed.

\subsection{Graph Evolution}
\label{sec:evolution}

A static skill graph cannot keep pace with a continuously improving policy: new failure modes demand new skills, redundant skills accumulate, and the relative importance of inter-skill relations shifts over training. To address this, we evolve both the skill nodes and their edges at each validation step, driven by trajectory-level feedback.

\subsubsection{Node-Level: Adaptive Granularity Control}
\label{sec:granularity}

We maintain appropriate skill granularity through four operations, each triggered by specific diagnostic signals from the training process.

\paragraph{Insert.} When the agent fails on tasks that existing skills do not adequately cover, we generate targeted new skills. The teacher model $\mathcal{M}$ analyzes a batch of failed trajectories $\tau^-$ together with the current skill set $\mathcal{R}_{\text{existing}}$, and proposes up to $m$ new skills addressing the identified failure causes:
\begin{equation}
    \{s_{\text{new}}^1, \ldots, s_{\text{new}}^m\} = \mathcal{M}(\text{insert},\, \tau^-,\, \mathcal{R}_{\text{existing}}).
    \label{eq:skill_insert}
\end{equation}

\paragraph{Merge.} Redundant skills inflate context length and dilute retrieval precision. We identify candidates for merging by measuring the overlap of their graph neighborhoods: let $\mathcal{N}(v)$ denote the set of neighbors of node $v$ in $\mathcal{G}$; when two skills $s_i$ and $s_j$ share most of their neighbors (Jaccard similarity $J(\mathcal{N}(s_i), \mathcal{N}(s_j)) \geq \tau_{\text{merge}}$, where $\tau_{\text{merge}}$ is the merge threshold), they likely encode redundant strategies and are synthesized into a single unified skill by $\mathcal{M}$.

\paragraph{Split.} Overly broad skills that conflate distinct sub-strategies exhibit moderate success rates despite high usage ($\hat{p}(v) \in [0.15, 0.4]$ and $n_{\text{use}}(v) \geq 10$). We decompose such skills into more focused sub-skills via $\mathcal{M}$, reconnecting them with prerequisite edges.

\paragraph{Deprecate.} Skills that are frequently retrieved but consistently fail ($n_{\text{use}}(v) \geq 20$ and $\hat{p}(v) < 0.15$) are deprecated and excluded from future retrieval, preventing them from degrading policy performance.

\subsubsection{Edge-Level: Topology Evolution}
\label{sec:edge_evolution}

While node-level operations adjust \emph{what} skills are available, edge-level operations adjust \emph{how} skills relate to one another, directly shaping retrieval quality.

\paragraph{Path reinforcement.} Successful trajectories provide evidence that the retrieved skill sequence was effective. We reinforce this signal by increasing the weight of every edge along the successful path:
\begin{equation}
    w(e) \leftarrow \min\bigl(w(e) + \alpha,\; 1.0\bigr), \quad \forall e \in \text{path}(\tau^+),
    \label{eq:path_reinforce}
\end{equation}
where $\alpha \in (0, 1)$ is the reinforcement step size and $\text{path}(\tau^+)$ denotes the set of edges traversed by the skill sequence used in successful trajectory $\tau^+$. This makes validated dependency paths more likely to be traversed in future retrieval.

\paragraph{Co-occurrence discovery.} New inter-skill relations emerge as the policy improves. When two skills co-occur in a successful episode but are not yet connected in $\mathcal{G}$, we add a $\texttt{co\_occur}$ edge to capture this discovered association.

\paragraph{Decay and pruning.} To prevent stale relations from persisting indefinitely, all edge weights undergo multiplicative decay with decay factor $\gamma \in (0, 1)$: $w(e) \leftarrow \gamma \cdot w(e)$ at each checkpoint. Edges whose weights fall below a pruning threshold $w_{\min}$ are removed from $\mathcal{E}$. After all updates, node levels $\ell(v)$ are recomputed to reflect the new topology.

\subsubsection{Progressive Unlocking}
\label{sec:unlocking}

Exposing the agent to complex, high-level skills before it has mastered their prerequisites can hinder learning. To implement a curriculum over skill complexity, \textsc{SkillGraph} progressively unlocks skills based on their topological level. Initially, only level-0 foundational skills are active. Let $L$ denote the current highest active level. When the average success rate of level-$L$ skills exceeds an unlocking threshold $\theta_{\text{unlock}}$, level-$(L{+}1)$ skills are activated:
\begin{equation}
    \bar{p}(L) = \frac{1}{|\{v : \ell(v) = L\}|} \sum_{v:\, \ell(v) = L} \hat{p}(v) \;\geq\; \theta_{\text{unlock}} \;\;\Longrightarrow\;\; \mathcal{V}_{\text{active}} \leftarrow \mathcal{V}_{\text{active}} \cup \{v : \ell(v) = L+1\}.
    \label{eq:unlock}
\end{equation}
This ensures that the agent builds competence from the ground up, with advanced compositional skills becoming available only when their foundations are reliable.

\subsection{Policy Optimization and Closed-Loop Training}
\label{sec:training}

We optimize the skill-augmented policy $\pi_\theta$, parameterized by $\theta$, using GRPO~\citep{shao2024deepseekmath}. For each task, we sample a group of $G$ rollouts from $\pi_\theta$ conditioned on the task description $d$ and the retrieved skill sequence $\mathcal{R}_{\text{ret}}$. Each rollout $i$ receives a binary reward $R_i \in \{0,1\}$ indicating task success, and the estimated advantage $\hat{A}_i$ is computed by within-group normalization:
\begin{equation}
    \hat{A}_i = \frac{R_i - \text{mean}(\{R_j\}_{j=1}^G)}{\text{std}(\{R_j\}_{j=1}^G) + \epsilon},
    \label{eq:advantage}
\end{equation}
where $\epsilon$ is a small constant for numerical stability. The policy is updated via the clipped surrogate objective with a KL penalty anchored to the reference policy $\pi_{\text{ref}}$ (initialized from the SFT model):
\begin{equation}
    \mathcal{L}(\theta) = \mathbb{E}\!\left[\min\!\left(r(\theta)\,\hat{A}_i,\; \text{clip}(r(\theta), 1\!-\!\epsilon_c, 1\!+\!\epsilon_c)\,\hat{A}_i\right) - \beta\, D_{\text{KL}}\!\left(\pi_\theta \,\|\, \pi_{\text{ref}}\right)\right],
    \label{eq:grpo}
\end{equation}
where $r(\theta) = \pi_\theta / \pi_{\text{old}}$ is the importance sampling ratio between the current and previous policies, $\epsilon_c$ is the clipping parameter, $\beta$ is the KL penalty coefficient, and $D_{\text{KL}}$ denotes the Kullback--Leibler divergence.

At each validation step, the full graph evolution pipeline is executed, creating a closed training loop: the improving policy generates richer trajectories that refine the skill graph through node- and edge-level updates, while the refined graph provides higher-quality structured retrieval that accelerates subsequent policy learning. The complete procedure is summarized in Algorithm~\ref{alg:skillgraph}.

\begin{algorithm}[t]
    \caption{\textsc{SkillGraph}: Skill-Augmented RL for Agents via Evolving Skill Graphs}
    \label{alg:skillgraph}
    \small
    \begin{algorithmic}[1]
    \REQUIRE Base policy $\pi_{\text{base}}$, teacher model $\mathcal{M}$, environment $\mathrm{Env}$, unlocking threshold $\theta_{\text{unlock}}$
    \ENSURE Trained policy $\pi_{\theta^*}$, evolved skill graph $\mathcal{G}^*$
    \STATE \textbf{--- Graph Construction ---}
    \STATE $\mathcal{T}^+, \mathcal{T}^- \leftarrow \text{Rollout}(\pi_{\text{base}}, \mathrm{Env})$
    \STATE $\mathcal{V} \leftarrow \mathcal{M}(\mathcal{T}^+, \mathcal{T}^-)$ \hfill $\triangleright$ Distill general \& task-specific skills
    \STATE $\mathcal{G}=(\mathcal{V},\mathcal{E}) \leftarrow \text{InitGraph}(\mathcal{V})$ \hfill $\triangleright$ Add prereq, enhance, co-occur edges
    \STATE Compute topological levels $\ell(v)$ for all $v \in \mathcal{V}$
    \STATE \textbf{--- Cold-Start SFT ---}
    \STATE $\theta \leftarrow \text{SFT}(\pi_{\text{base}},\, \mathcal{M}(\mathrm{Env}, \mathcal{G}))$; \quad $\pi_{\text{ref}} \leftarrow \pi_\theta$
    \STATE $\mathcal{V}_{\text{active}} \leftarrow \{v : \ell(v) = 0\}$; \quad $L \leftarrow 0$ \hfill $\triangleright$ Unlock level-0 skills
    \STATE \textbf{--- Closed-Loop RL Training ---}
    \FOR{epoch $= 1$ to $N$}
        \FOR{each task $d$ with type $t(d)$}
            \STATE \textbf{Graph-Aware Retrieval:}
            \STATE \quad $\mathcal{R}_{\text{seed}} \leftarrow \{v \in \mathcal{V}_{\text{active}} : \mathrm{category}(v)=\texttt{general} \vee \mathrm{category}(v)=t(d)\}$
            \STATE \quad $\mathcal{R}_{\text{BFS}} \leftarrow \text{BackwardBFS}(\mathcal{R}_{\text{seed}}, \mathcal{G}, D)$; \quad $\mathcal{R}_{\text{beam}} \leftarrow \text{ForwardBeam}(\mathcal{R}_{\text{seed}}, \mathcal{G}, B)$
            \STATE \quad $\mathcal{R}_{\text{ret}} \leftarrow \text{TopoSort}_{\mathcal{G}}(\mathcal{R}_{\text{seed}} \cup \mathcal{R}_{\text{BFS}} \cup \mathcal{R}_{\text{beam}})$ \hfill $\triangleright$ Cap at $K_{\max}$ skills
            \STATE Sample $G$ rollouts $\{\tau^{(i)}\}_{i=1}^G \sim \pi_\theta(\cdot \mid d, \mathcal{R}_{\text{ret}})$; \quad Update $\theta$ via GRPO
        \ENDFOR
        \IF{validation step}
            \STATE \textbf{Graph Evolution:}
            \STATE \textit{Node-level:} Insert / Merge / Split / Deprecate skills via $\mathcal{M}$
            \STATE \textit{Edge-level:} Reinforce paths in $\tau^+$; Discover new co-occur edges; Decay \& prune weak edges
            \STATE Recompute topological levels $\ell(v)$
            \STATE \textbf{Progressive Unlocking:} \textbf{if} $\bar{p}(L) \geq \theta_{\text{unlock}}$ \textbf{then} $\mathcal{V}_{\text{active}} \leftarrow \mathcal{V}_{\text{active}} \cup \{v : \ell(v) = L{+}1\}$; $L \leftarrow L{+}1$
        \ENDIF
    \ENDFOR
    \RETURN $\pi_\theta, \mathcal{G}$
    \end{algorithmic}
    \end{algorithm}
\vspace{10pt}
\section{Experiments}
\label{sec:experiments}
\subsection{Experimental Setup}

\begin{table}[t]
    \caption{Main results on ALFWorld and WebShop. ALFWorld reports per-subtask and overall success rates~(\%); WebShop reports task score and success rate~(\%). \textbf{Bold} and \underline{underline} denote the best and second-best results, respectively.}
    \label{tab:main}
    \centering
    \small
    \setlength{\tabcolsep}{4pt}
    \begin{tabular}{lrrrrrrr|rr}
    \toprule
    \multirow{2}{*}{Method} & \multicolumn{7}{c|}{ALFWorld} & \multicolumn{2}{c}{WebShop} \\
     & Pick & Look & Clean & Heat & Cool & Pick2 & All & Score & Succ. \\
    \midrule
    \multicolumn{10}{l}{\textit{Closed-source LLMs}} \\
    GPT-4o          & 75.3 & 60.8 & 31.2 & 56.7 & 21.6 & 49.8 & 48.0 & 31.8 & 23.7 \\
    Gemini-2.5-Pro  & 92.8 & 63.3 & 62.1 & 69.0 & 26.6 & 58.7 & 60.3 & 42.5 & 35.9 \\
    \midrule
    \multicolumn{10}{l}{\textit{Prompt-based Agentic or Memory-based Methods}} \\
    ReAct     & 48.5 & 35.4 & 34.3 & 13.2 & 18.2 & 17.6 & 31.2 & 46.2 & 19.5 \\
    Reflexion  & 62.0 & 41.6 & 44.9 & 30.9 & 36.3 & 23.8 & 42.7 & 58.1 & 28.8 \\
    Mem0            & 54.0 & 55.0 & 26.9 & 36.4 & 20.8 & 7.69 & 33.6 & 23.9 & 2.00 \\
    MemP            & 54.3 & 38.5 & 48.1 & 56.2 & 32.0 & 16.7 & 41.4 & 25.3 & 6.40 \\
    ExpeL           & 21.0 & 67.0 & 55.0 & 52.0 & 11.0 & 6.00 & 46.3 & 30.9 & 11.2 \\
    SimpleMem       & 64.5 & 33.3 & 20.0 & 12.5 & 33.3 & 3.84 & 29.7 & 33.2 & 8.59 \\
    \midrule
    \multicolumn{10}{l}{\textit{RL-based Methods}} \\
    RLOO       & 87.6 & \underline{78.2} & 87.3 & 81.3 & 71.9 & 48.9 & 75.5 & 80.3 & 65.7 \\
    GRPO       & 90.8 & 66.1 & 89.3 & 74.7 & 72.5 & 64.7 & 77.6 & 79.3 & 66.1 \\
    \midrule
    \multicolumn{10}{l}{\textit{Memory-Augmented RL-based Methods}} \\
    MemRL           & 62.8 & 38.5 & 22.2 & 12.5 &  8.00 &  0.00 & 21.4 & 29.5 &  9.20 \\
    EvolveR         & 64.9 & 33.3 & 46.4 & 13.3 & 33.3 & 33.3 & 43.8 & 42.5 & 17.6 \\
    Mem0+GRPO       & 78.1 & 54.8 & 56.1 & 31.0 & 65.0 & 26.9 & 54.7 & 58.1 & 37.5 \\
    SimpleMem+GRPO  & 89.5 & 63.6 & 60.0 & 50.0 & 64.9 & 26.3 & 62.5 & 67.8 & 46.9 \\
    SkillRL         & \underline{97.9} & 71.4 & \underline{90.0} & \underline{90.0} & \textbf{95.5} & \textbf{87.5} & \underline{89.9} & \underline{85.2} & \underline{72.7} \\
    \midrule
    \textsc{SkillGraph} (Ours) & \textbf{100.0} & \textbf{80.0} & \textbf{100.0} & \textbf{100.0} & \underline{95.3} & \underline{83.3} & \textbf{93.1} & \textbf{91.5} & \textbf{84.4} \\
    \bottomrule
    \end{tabular}
    \end{table}
    
\paragraph{Environments.}
ALFWorld~\citep{shridhar2020alfworld} is a text-based household interaction environment that covers six task categories (Pick, Look, Clean, Heat, Cool, Pick2), each requiring multi-step goal-directed manipulation. WebShop~\citep{yao2022webshop} presents a web navigation challenge in which agents must search, browse, and purchase products meeting specific user requirements. For search-augmented question answering, we evaluate on three single-hop benchmarks---NQ~\citep{kwiatkowski2019nq}, TriviaQA~\citep{joshi2017triviaqa}, and PopQA~\citep{mallen2023popqa}---and four multi-hop benchmarks---HotpotQA~\citep{yang2018hotpotqa}, 2Wiki~\citep{ho2020twowiki}, MuSiQue~\citep{trivedi2022musique}, and Bamboogle~\citep{press2023bamboogle}.

\paragraph{Baselines.}
We compare \textsc{SkillGraph} against four groups of methods. \textbf{(1) Closed-source LLMs}: GPT-4o~\citep{openai2024gpt4o} and Gemini-2.5-Pro~\citep{comanici2025gemini}, serving as strong references. \textbf{(2) Prompt-based and memory-augmented methods}: ReAct~\citep{yao2022react}, Reflexion~\citep{shinn2023reflexion}, Mem0~\citep{chhikara2025mem0}, MemP~\citep{fang2025memp}, ExpeL~\citep{zhao2024expel}, and SimpleMem~\citep{liu2025simplemem}, which use in-context experience without parameter updates. \textbf{(3) RL-based methods}: RLOO~\citep{ahmadian2024rloo} and GRPO~\citep{shao2024deepseekmath}. \textbf{(4) Memory-augmented RL methods}: MemRL~\citep{zhang2026memrl}, EvolveR~\citep{wu2025evolver}, Mem0+GRPO, SimpleMem+GRPO, and SkillRL~\citep{xia2025skillrl}. For search-augmented QA, we additionally compare against CoT~\citep{wei2022chain}, RAG~\citep{arslan2024survey}, Search-o1~\citep{li2025searcho1}, Search-R1~\citep{jin2025searchr1} and ZeroSearch~\citep{sun2025zerosearch}.

\paragraph{Implementation details.}
We adopt Qwen2.5-7B-Instruct~\citep{bai2023qwen} as the base policy $\pi_{\text{base}}$, initialized via cold-start SFT, and OpenAI o3~\citep{openai2025o3} as the teacher model $\mathcal{M}$ for skill distillation, SFT data generation, and graph evolution operations. RL training uses GRPO with learning rate $1 \times 10^{-6}$, KL coefficient $\beta=0.01$, clipping parameter $\epsilon_c=0.2$, train batch size $16$, and group size $G=8$. For graph-aware retrieval, we cap the retrieved skill sequence at $K_{\max}=8$, set backward-BFS depth $D=2$, and forward beam width $B=3$. For graph evolution, edges are initialized with weights $w=0.3$ (co-occur) and $w=0.2$ (enhance); at each validation checkpoint, successful paths receive additive reinforcement $\alpha=0.05$, all weights decay by factor $\gamma=0.99$, and edges below $w_{\min}=0.05$ are pruned. Node-level evolution uses merge threshold $\tau_{\text{merge}}=0.85$, and at most $m=3$ newly inserted skills per update. Progressive unlocking activates level-$(L{+}1)$ skills when the average success rate of level-$L$ skills exceeds $\theta_{\text{unlock}}=0.6$.

\subsection{Main Results}

\paragraph{Comparison with baselines.}
Table~\ref{tab:main} reports results on ALFWorld and WebShop. \textsc{SkillGraph} achieves the best overall performance on both benchmarks.
\textit{(i)} Notably, \textsc{SkillGraph} with a 7B open-source model substantially outperforms closed-source LLMs: it surpasses GPT-4o by $45.1$ points and Gemini-2.5-Pro by $32.8$ points on ALFWorld, and exceeds both by over $48$ points on WebShop, demonstrating that structured skill reasoning can compensate for the scale gap.
\textit{(ii)} Compared with prompt-based and memory methods, \textsc{SkillGraph} outperforms the best method (ExpeL) by $46.8$ points on ALFWorld, with the largest gains on Clean ($100.0$ vs.\ $55.0$) and Heat ($100.0$ vs.\ $52.0$). These subtasks require executing prerequisite actions in a strict order, which flat retrieval cannot enforce but graph-aware retrieval handles naturally.
\textit{(iii)} Over the vanilla GRPO baseline with the same optimizer, \textsc{SkillGraph} improves by $15.5$ and $18.3$ points on ALFWorld and WebShop respectively, directly quantifying the benefit of graph-structured skill guidance in reducing exploration burden.
\textit{(iv)} Against the strongest prior method SkillRL, \textsc{SkillGraph} achieves slightly higher ALFWorld performance while gaining $11.7$ points on WebShop. The gap stems from the evolving graph structure: graph evolution continuously refines the skill set and discovers inter-skill relations (e.g., query refinement $\to$ attribute matching $\to$ price comparison), providing higher-quality compositional guidance than a static flat skill bank.

\paragraph{Generalization to search-augmented QA.}

\begin{table}[t]
    \caption{Results on search-augmented QA. \textsc{SkillGraph} is trained on NQ$^\heartsuit$ and HotpotQA$^\heartsuit$ (in-domain) and evaluated zero-shot on the remaining five benchmarks$^\spadesuit$ (out-of-domain).}
    \label{tab:qa}
    \centering
    \footnotesize
    \setlength{\tabcolsep}{3pt}
    \begin{tabular}{lrrr|rrrr|r}
    \toprule
    \multirow{2}{*}{Method} & \multicolumn{3}{c|}{Single-Hop QA} & \multicolumn{4}{c|}{Multi-Hop QA} & \multirow{2}{*}{Avg.} \\
     & NQ$^\heartsuit$ & TriviaQA$^\spadesuit$ & PopQA$^\spadesuit$ & HotpotQA$^\heartsuit$ & 2Wiki$^\spadesuit$ & MuSiQue$^\spadesuit$ & Bamboogle$^\spadesuit$ & \\
    \midrule
    Qwen2.5-7B-Instruct  & 11.6 & 35.6 &  1.20 & 16.4 & 22.2 &  4.80 & 14.4 & 15.2 \\
    CoT   & 12.8 & 35.6 &  3.80 & 16.2 & 22.6 &  6.60 & 24.0 & 17.4 \\
    RAG      & 27.4 & 58.2 & 17.8  & 25.8 & 23.2 &  9.40 & 16.8 & 25.5 \\
    Search-o1 & 19.4 & 40.6 & 11.4  & 17.0 & 27.0 &  8.60 & 30.4 & 22.1 \\
    Search-R1         & 39.3 & 61.0 & 39.7  & 37.0 & 40.1 & 14.6  & 36.8 & 38.5 \\
    ZeroSearch        & 43.6 & 61.8 & \textbf{51.5}  & 34.6 & 35.2 & 18.4  & 27.8 & 39.1 \\
    EvolveR           & 43.5 & \underline{63.4} & 44.6  & 38.2 & \underline{42.0} & 15.6  & 54.4 & 43.1 \\
    SkillRL           & \underline{45.9} & 63.3 & 45.9  & \underline{43.2} & 40.3 & \textbf{20.2}  & \textbf{73.8} & \underline{47.1} \\
    \midrule
    \textsc{SkillGraph} (Ours) & \textbf{48.0} & \textbf{63.8} & \underline{48.5} & \textbf{44.7} & \textbf{43.4} & \underline{19.5} & \underline{72.6} & \textbf{48.9} \\
    \bottomrule
    \end{tabular}
    \end{table}
Table~\ref{tab:qa} reports results on seven QA benchmarks. Trained only on NQ and HotpotQA, \textsc{SkillGraph} achieves the highest average performance ($48.6$) and generalizes zero-shot to five unseen datasets.
On single-hop tasks, \textsc{SkillGraph} surpasses all baselines on NQ ($48.0$) and TriviaQA ($63.8$), improving over SkillRL by $2.1$ and $0.5$ respectively. This advantage stems from graph evolution, which keeps the skill set aligned with the evolving policy rather than relying on a fixed skill library.
On multi-hop tasks, \textsc{SkillGraph} leads on HotpotQA ($44.7$) and 2Wiki ($43.4$), where prerequisite-ordered retrieval helps decompose chained queries into sub-questions.
These results confirm that the structured skill representation learned from two training domains transfers effectively to unseen tasks, demonstrating strong generalization without task-specific adaptation.

\begin{figure}[t]
    \centering
    \includegraphics[width=\linewidth]{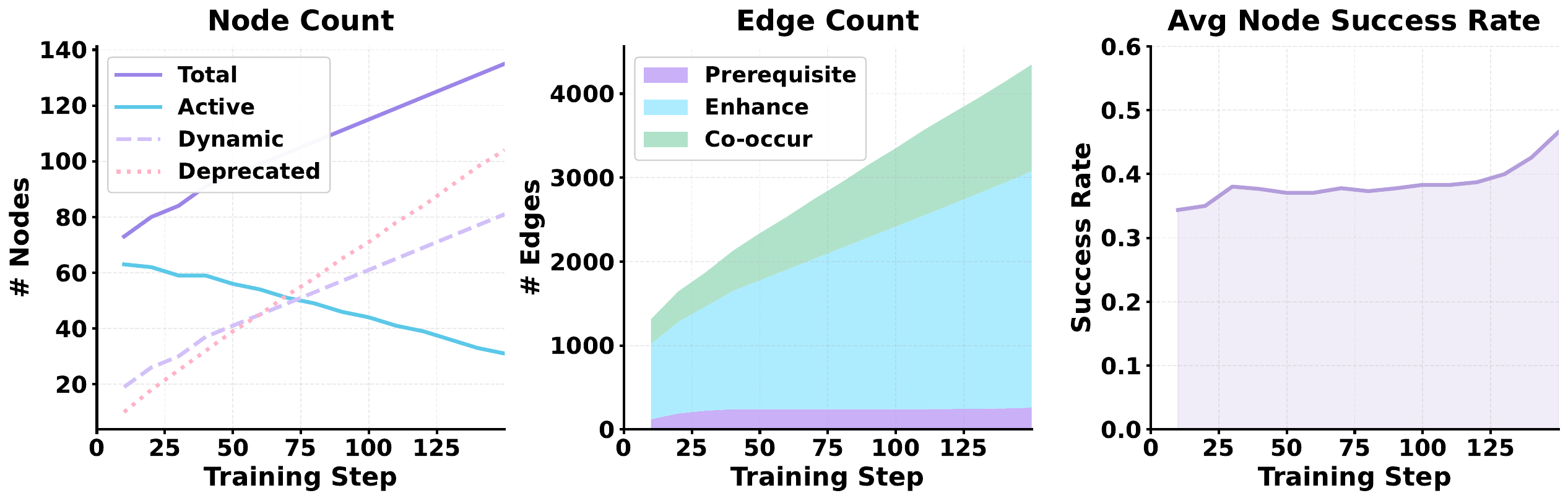}
    \vspace{-20pt}
    \caption{Skill graph evolution over training on WebShop. Left: node counts (total, active, inserted, deprecated). Middle: edge counts by type. Right: average node success rate.}
    \label{fig:graph_evolution}
\end{figure}
\subsection{Analysis}

\begin{wraptable}{l}{0.5\textwidth}
    \vspace{-20pt}
    \caption{Ablation study on ALFWorld and WebShop success rate(\%).}
    \label{tab:ablation}
    \centering
    \small
    \begin{tabular}{lrr}
    \toprule
    Method & ALFWorld & WebShop \\
    \midrule
    \textsc{SkillGraph} & \textbf{93.1} & \textbf{84.4} \\
    \midrule
    w/o Graph Structure  & 89.9 & 72.7 \\
    w/o Graph-aware Ret. &  59.4    &  79.7  \\
    w/o Graph Evolution &  78.2  &  70.3 \\
    w/o Cold-Start SFT   &  73.4  &  67.2  \\
    \bottomrule
    \end{tabular}
    \vspace{-20pt}
\end{wraptable}
\paragraph{Ablation study.}
Table~\ref{tab:ablation} isolates each component's contribution. The components exhibit complementary strengths across environments.
On ALFWorld, removing graph-aware retrieval causes the largest single drop ($-33.7$), confirming that the rigid multi-step subtasks (e.g., Clean, Heat) critically depend on prerequisite-ordered skill sequences, consistent with the large gains reported in the main results.
On WebShop, graph evolution ($-14.1$) and graph structure ($-11.7$) contribute the most, indicating that WebShop benefits primarily from maintaining a high-quality, evolving skill set---the correct skills matter more than their ordering in this flexible navigation setting, which explains why the graph structure gap over SkillRL ($+11.7$) is larger than the retrieval ordering gap.
Cold-start SFT yields a large combined drop, confirming that a good initialization is essential for RL convergence in complex agent environments.

\paragraph{Skill graph evolution dynamics.}
Figure~\ref{fig:graph_evolution} tracks graph statistics over training. Node count grows from ${\sim}20$ to ${\sim}140$ via failure-driven insertion, but the active count plateaus earlier as deprecation prunes failing skills---a self-regulating loop that prevents unbounded growth. Co-occur edges grow fastest through automatic discovery, while prerequisite and enhance edges increase steadily via path reinforcement, showing that the graph discovers relational structure beyond initial construction. The average node success rate rises confirming that evolution progressively filters low-quality skills while reinforcing useful ones.

\begin{figure}[t]
    \centering
    \begin{minipage}[t]{0.48\linewidth}
        \centering
        \includegraphics[width=\linewidth]{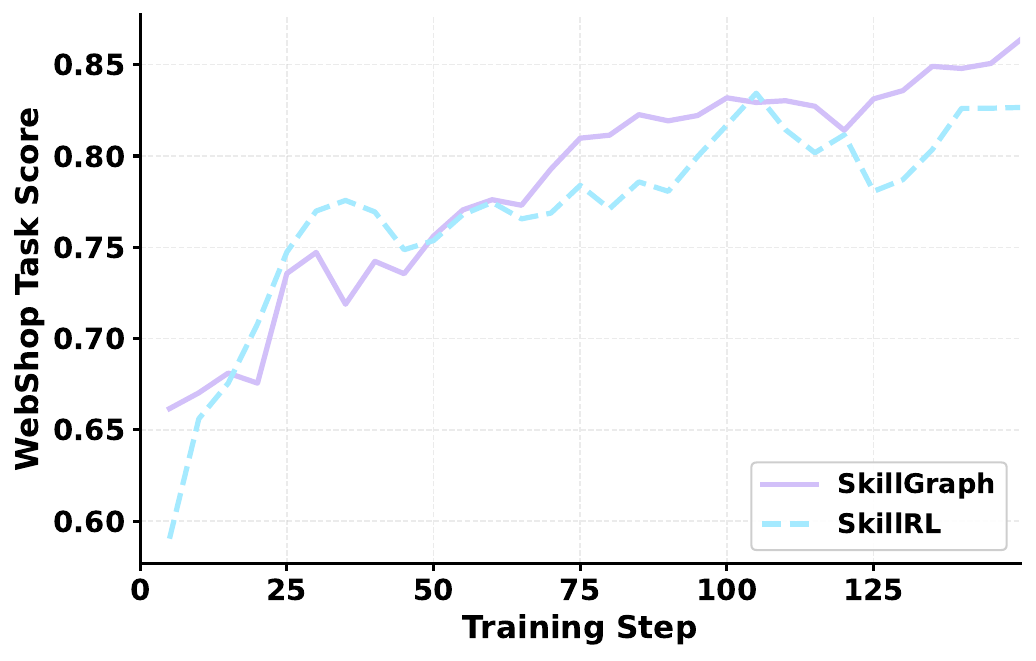}
    \end{minipage}\hfill
    \begin{minipage}[t]{0.48\linewidth}
        \centering
        \includegraphics[width=\linewidth]{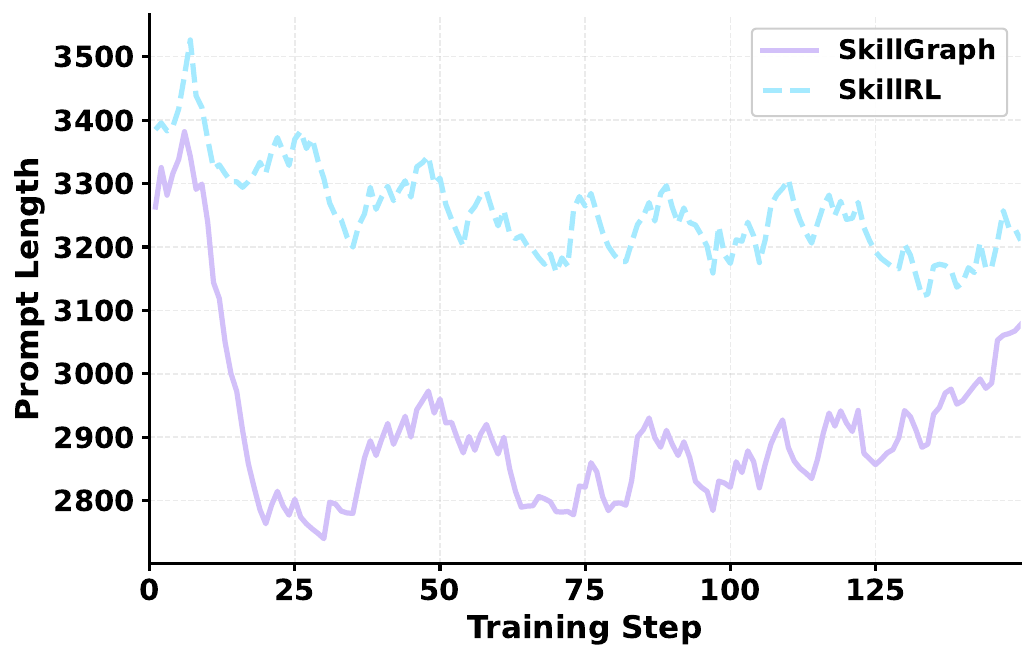}
    \end{minipage}
    \caption{Training dynamics and context efficiency. Left: WebShop task score over training epochs. Right: average prompt length during training.}
    \label{fig:analysis_curves}
\end{figure}

\paragraph{Convergence and context efficiency.}
Figure~\ref{fig:analysis_curves}~(left) shows that \textsc{SkillGraph} surpasses SkillRL after roughly $50$ training steps and maintains a consistently higher score thereafter, converging to a superior final performance. The faster convergence is driven by dependency-ordered retrieval reducing early-stage exploration and progressive unlocking acting as an automatic curriculum. Figure~\ref{fig:analysis_curves}~(right) shows that graph-guided retrieval maintains shorter prompts than flat retrieval throughout training, because graph traversal limits the retrieved set to topologically relevant skills rather than all semantically similar entries, improving both inference cost and signal-to-noise ratio.

\section{Conclusion}
\label{sec:conclusion}

We presented \textsc{SkillGraph}, a framework that organizes agent skills into a structured dependency graph with typed relational edges and co-evolves the graph with the agent's policy through RL. By unifying graph construction, graph-aware retrieval, and graph evolution into a closed training loop, \textsc{SkillGraph} addresses three key limitations of flat skill libraries: weak compositional planning, poor granularity control, and the inability to accumulate inter-skill relational signals. Experiments on ALFWorld, WebShop, and seven search-augmented QA benchmarks demonstrate state-of-the-art performance, with the largest gains on complex multi-step tasks that require ordered skill composition.

\paragraph{Limitations and future work.}
Our current framework relies on a strong teacher model for skill distillation and graph-adaptive operations, which introduces additional inference cost during graph evolution. Exploring lightweight alternatives such as self-distillation or critic-based skill generation could reduce this dependency. Additionally, the skill graph is currently constructed and evolved within a single environment; investigating cross-environment skill transfer---where a graph trained on one domain bootstraps learning in another---is a promising direction. Finally, scaling \textsc{SkillGraph} to larger base models and more diverse task distributions remains an open question worth exploring.

\clearpage
\subsection*{AI use statement}
Large language models were used in this work in two capacities. \textbf{(1)~As part of the research methodology:} LLMs serve as the teacher model for skill distillation, SFT data generation, and graph evolution operations (insertion, merge, split), and as the base policy fine-tuned via RL, as described in Section~\ref{sec:method}. \textbf{(2)~For writing assistance:} LLMs were used to polish the language and improve the presentation of this manuscript. All LLM-assisted content has been manually reviewed, verified, and edited by the authors. The authors take full responsibility for the accuracy and integrity of all claims, results, and statements presented in this paper.

\subsection*{Ethics statement}
This work evaluates agents in simulated household tasks, web navigation, and search-augmented question answering. Skills distilled from trajectories may inherit biases or errors from the underlying language models and teacher-generated guidance. Applications beyond the evaluated environments require careful validation and appropriate safeguards. Appendix~\ref{app:broader_impact} discusses potential benefits, risks, and limitations, and Appendix~\ref{app:licenses} describes the assets used in this work.

\subsection*{Reproducibility statement}
Section~\ref{sec:method} and Algorithm~\ref{alg:skillgraph} describe the proposed framework. Appendices~\ref{app:entire_pipeline} and~\ref{app:implementation} provide graph construction, evolution, and implementation details. Appendix~\ref{app:experimental_details} lists training and graph hyperparameters, and Appendix~\ref{app:prompts} provides representative prompt templates.

\bibliographystyle{iclr2027_conference}
\bibliography{references}
\appendix
\section{Supplementary Details for \textsc{SkillGraph}}
\label{app:entire_pipeline}

This appendix provides formal definitions, derivations, and implementation specifics that supplement the method description in Section~\ref{sec:method}.

\subsection{Level Computation}
\label{app:level_computation}

Each node $v \in \mathcal{V}$ is assigned a topological level $\ell(v)$ used for progressive unlocking and dependency-respecting retrieval ordering. Levels are computed via BFS over the directional dependency edges:
\begin{equation}
    \ell(v) = \begin{cases} 0 & \text{if } v \text{ has no prerequisite/enhancement parents}, \\ \max_{u: (u,v) \in \mathcal{E}_{\text{dep}}} \ell(u) + 1 & \text{otherwise}, \end{cases}
    \label{eq:level}
\end{equation}
where $\mathcal{E}_{\text{dep}} = \{e \in \mathcal{E} : \text{type}(e) \in \{\texttt{prereq}, \texttt{enhance}\}\}$. We exclude $\texttt{co\_occur}$ edges from this computation because co-occurrence captures symmetric association rather than directional dependency; including them would introduce cycles and blur the prerequisite hierarchy. Levels are recomputed after every graph evolution step to reflect topology changes.

\subsection{Edge Initialization}
\label{app:edge_init}

Before training begins, edges are initialized with structural priors rather than learned from data: $\texttt{co\_occur}$ edges ($w=0.3$) connect task-specific skills within the same category, and $\texttt{enhance}$ edges ($w=0.2$) connect each general skill to all task-specific skills. No $\texttt{prereq}$ edges are created at initialization; they emerge through graph evolution as the agent discovers ordering dependencies.

\subsection{Statistics Update}
\label{app:stats_update}

At each validation checkpoint, node statistics are updated incrementally before evolution decisions are made:
\begin{equation}
    \hat{p}(v) \leftarrow \frac{n_{\text{succ}}(v) + n_{\text{succ}}^{\text{new}}(v)}{n_{\text{use}}(v) + n_{\text{use}}^{\text{new}}(v)},
    \label{eq:stats_update}
\end{equation}
where $n_{\text{succ}}^{\text{new}}(v)$ and $n_{\text{use}}^{\text{new}}(v)$ are the success and usage counts observed in the latest trajectory batch. A skill receives one usage count when it is retrieved into the prompt and one success count when the corresponding rollout succeeds.

\subsection{Node Evolution: Additional Details}
\label{app:node_evolution}

\paragraph{Insertion and edge bootstrapping.} Newly inserted skills start as isolated nodes with no edges. Connections are established in subsequent checkpoints via the co-occurrence discovery mechanism: if a new skill and an existing skill co-appear in at least $c_{\min}=2$ successful episodes, a $\texttt{co\_occur}$ edge is added automatically.

\paragraph{Merge: edge inheritance.} When two skills $v_i$ and $v_j$ are merged into $v_{\text{merged}}$, the merged node inherits the union of edges from both originals. Duplicate edges to the same neighbor are resolved by keeping the higher weight.

\paragraph{Split: edge reconnection.} When a skill $v$ is split into sub-skills $\{v_1', v_2', \ldots\}$, the sub-skills are connected by $\texttt{prereq}$ edges in the order produced by the teacher model. Existing edges of $v$ are redistributed to the sub-skill whose description is most relevant.

\subsection{Progressive Unlocking: Additional Details}
\label{app:unlocking_details}

During the initial warmup phase (the first 5 training steps), only non-deprecated level-0 skills are active:
$\mathcal{V}_{\text{active}}^{(0)} = \{v : \ell(v) = 0\}$.
After warmup, unlocking is checked at each validation checkpoint. Success rates are smoothed with a Beta$(1,1)$ prior to avoid premature unlocking from small sample sizes. If the newly unlocked level already satisfies the threshold, multiple levels can be unlocked within a single checkpoint, enabling rapid progression when the policy has strong foundational competence.

\section{Additional Implementation Details}
\label{app:implementation}

\paragraph{Skill schema.}
Each natural-language skill is stored as a compact record containing a unique skill identifier, a short title, a principle, an applicability condition, and a category (\texttt{general} or an environment-specific task type). The same record format is used by flat skill-library baselines; \textsc{SkillGraph} augments it with graph metadata including level, exposure count, successful-exposure count, success rate, creation step, and deprecation status.

\paragraph{Co-occurrence edge threshold.}
New $\texttt{co\_occur}$ edges require at least $c_{\min}=2$ co-appearances in successful validation episodes before being added, preventing spurious edges from single lucky episodes. Deprecated nodes are retained in the saved graph for auditability but excluded from $\mathcal{V}_{\text{active}}$.

\section{Experimental Details}
\label{app:experimental_details}

\paragraph{Metric definitions.}
For ALFWorld, we report success rate. For WebShop, Table~\ref{tab:main} reports both normalized task score and binary success rate, while Table~\ref{tab:ablation} reports task score to match the training-curve analysis. For QA benchmarks, we report exact-match accuracy under the search-augmented QA evaluation protocol. The search experiments use a tool-augmented QA environment where the retriever returns top-$3$ passages from a Wikipedia index built with an E5 retriever. \textsc{SkillGraph} is trained on NQ and HotpotQA and evaluated on the seven datasets reported in Table~\ref{tab:qa}.

\paragraph{Hyperparameters.}
Table~\ref{tab:hyperparams_training} lists the training hyperparameters for each environment. Table~\ref{tab:hyperparams_graph} lists the \textsc{SkillGraph}-specific hyperparameters, which are shared across all three environments.

\begin{table}[h]
    \caption{Training hyperparameters per environment.}
    \label{tab:hyperparams_training}
    \centering
    \footnotesize
    \setlength{\tabcolsep}{4pt}
    \begin{tabular}{lccc}
    \toprule
    Hyperparameter & ALFWorld & WebShop & Search QA \\
    \midrule
    \multicolumn{4}{l}{\textit{Optimization}} \\
    Learning rate & $1\!\times\!10^{-6}$ & $1\!\times\!10^{-6}$ & $1\!\times\!10^{-6}$ \\
    LR warmup ratio & -- & -- & 0.1 \\
    KL coefficient $\beta$ & 0.01 & 0.01 & 0.001 \\
    Entropy coefficient & -- & -- & 0 \\
    Invalid-action penalty & 0.1 & 0.1 & 0.01 \\
    \midrule
    \multicolumn{4}{l}{\textit{Batch sizes}} \\
    Train batch size & 16 & 16 & 256 \\
    Validation batch size & 64 & 64 & 512 \\
    Group size $G$ & 8 & 8 & 4 \\
    PPO mini-batch size & 128 & 64 & 256 \\
    PPO micro-batch / GPU & 4 & 4 & 8 \\
    \midrule
    \multicolumn{4}{l}{\textit{Sequence lengths}} \\
    Max prompt length & 4096 & 6000 & 5000 \\
    Max response length & 512 & 768 & 700 \\
    Max episode steps & 50 & 15 & 4 \\
    \midrule
    \multicolumn{4}{l}{\textit{Infrastructure}} \\
    Tensor-parallel size & 4 & 4 & 1 \\
    GPU memory utilization & 0.5 & 0.7 & 0.5 \\
    Param offload & \checkmark & \checkmark & -- \\
    Optimizer offload & \checkmark & \checkmark & -- \\
    \midrule
    \multicolumn{4}{l}{\textit{Schedule}} \\
    Total epochs & 200 & 200 & 200 \\
    Save frequency & 10 & 10 & 5 \\
    Validation frequency & 5 & 5 & 5 \\
    \bottomrule
    \end{tabular}
\end{table}

\begin{table}[h]
    \caption{\textsc{SkillGraph} hyperparameters (shared across all environments).}
    \label{tab:hyperparams_graph}
    \centering
    \footnotesize
    \begin{tabular}{lll}
    \toprule
    Hyperparameter & Symbol & Value \\
    \midrule
    \multicolumn{3}{l}{\textit{Graph-aware retrieval}} \\
    Retrieved skill cap & $K_{\max}$ & 8 \\
    Backward BFS depth & $D$ & 2 \\
    Forward beam width & $B$ & 3 \\
    \midrule
    \multicolumn{3}{l}{\textit{Node-level evolution}} \\
    Max new skills per update & $m$ & 3 \\
    Merge threshold (Jaccard) & $\tau_{\text{merge}}$ & 0.85 \\
    Deprecation threshold & -- & 0.15 \\
    Min usage for deprecation & -- & 20 \\
    \midrule
    \multicolumn{3}{l}{\textit{Edge-level evolution}} \\
    Path reinforcement step & $\alpha$ & 0.05 \\
    Edge decay factor & $\gamma$ & 0.99 \\
    Edge pruning threshold & $w_{\min}$ & 0.05 \\
    \midrule
    \multicolumn{3}{l}{\textit{Progressive unlocking}} \\
    Curriculum warmup epochs & -- & 5 \\
    Level unlock threshold & $\theta_{\text{unlock}}$ & 0.6 \\
    \bottomrule
    \end{tabular}
\end{table}

\section{Prompt Templates}
\label{app:prompts}

This section gives representative prompt templates used by \textsc{SkillGraph}. Environment-specific prompts differ mainly in the action space and observation format, while the retrieved skill block is shared across ALFWorld, WebShop, and search-augmented QA.

\paragraph{Agent prompt with retrieved skills.}
The following template shows the common structure used when skill memory is enabled. The ALFWorld and WebShop variants replace the action-space description with admissible environment actions, while the search variant replaces it with the choice between issuing a \texttt{<search>} query and returning a \texttt{<answer>}.

\begin{tcolorbox}[colback=gray!5, colframe=gray!60, title={\small\textbf{Agent Prompt Template}}, fontupper=\small\ttfamily, breakable, left=4pt, right=4pt, top=2pt, bottom=2pt]
You are an expert agent operating in the target environment.\\
Your task is to: \{task\_description\}\\[4pt]
\#\# Retrieved Relevant Experience\\[2pt]
\{retrieved\_skills\}\\[4pt]
\#\# Current Progress\\[2pt]
Prior to this step, you have already taken \{step\_count\} step(s).\\
Below are the most recent observations and actions:\\
\{action\_history\}\\[4pt]
Current observation:\\
\{current\_observation\}\\[4pt]
Admissible actions:\\
{[}\{admissible\_actions\}{]}\\[4pt]
Now it is your turn to take an action.\\
First reason step-by-step within <think> </think> tags.\\
Then choose one admissible action within <action> </action> tags.
\end{tcolorbox}

\paragraph{Graph-ordered skill injection.}
In graph retrieval mode, retrieved skills are rendered in dependency order before being inserted into the agent prompt. This makes the graph structure visible to the policy without requiring special model architecture changes.

\begin{tcolorbox}[colback=blue!3, colframe=blue!40, title={\small\textbf{Graph-Ordered Skill Block}}, fontupper=\small\ttfamily, breakable, left=4pt, right=4pt, top=2pt, bottom=2pt]
\#\#\# Skills (ordered by dependency)\\
- **[category] Skill Title** [skill\_id]: Skill principle.\\
\quad\_Apply when: Applicability condition.\_\\
- **Next Skill Title** [skill\_id]: Skill principle.\\
\quad\_Apply when: Applicability condition.\_\\[4pt]
\#\#\# Mistakes to Avoid\\
- **Don't**: Failure pattern or bad action.\\
\quad**Instead**: Corrective strategy.
\end{tcolorbox}

\paragraph{Failure-driven skill insertion prompt.}
During dynamic updates, failed validation trajectories are summarized and passed to the teacher model. The teacher is asked to produce a small number of new skills and to avoid duplicating existing skill titles. Returned identifiers are reassigned by the implementation to prevent collisions.

\begin{tcolorbox}[colback=orange!3, colframe=orange!50, title={\small\textbf{Failure-Driven Skill Insertion Prompt}}, fontupper=\small\ttfamily, breakable, left=4pt, right=4pt, top=2pt, bottom=2pt]
Analyze these failed agent trajectories and suggest NEW skills\\
to add to the skill bank.\\[4pt]
FAILED TRAJECTORIES:\\
Example 1:\\
\quad Task: \{task\}\\
\quad Task Type: \{task\_type\}\\
\quad Trajectory (last 5 steps):\\
\quad\quad Action: \{action\}\\
\quad\quad Observation: \{observation\}\\
\quad ...\\[4pt]
EXISTING SKILL TITLES (avoid duplicating these):\\
\{existing\_titles\}\\[4pt]
Generate 1-\{max\_new\_skills\} NEW actionable skills that would\\
help avoid these failures.\\
Each skill must have: skill\_id, title (3-5 words), principle\\
(1-2 sentences), when\_to\_apply.\\[4pt]
Use skill\_ids: \{dyn\_id\_list\}\\[4pt]
Return ONLY a JSON array of skills, no other text.
\end{tcolorbox}

\paragraph{Skill merge and split prompts.}
For graph evolution, the teacher is also used as a skill-bank curator. Merge prompts ask it to combine two semantically overlapping skills into one concise skill. Split prompts ask it to decompose a high-usage but low-success skill into two or three simpler sub-skills, optionally conditioned on failure contexts where the original skill did not help. Both prompts require the same JSON schema as insertion: \texttt{skill\_id}, \texttt{title}, \texttt{principle}, and \texttt{when\_to\_apply}.

\section{Additional Search Training Results}
\label{app:search_curve}

Table~\ref{tab:search_curve} reports intermediate validation checkpoints for the search-augmented QA \textsc{SkillGraph} run. The final checkpoint at step $200$ gives the best average score, while NQ and HotpotQA peak slightly earlier at step $180$. We report the unified step-$200$ checkpoint in Table~\ref{tab:qa} for a single consistent model selection rule across datasets.

\begin{table}[h]
    \caption{Search-augmented QA validation accuracy (\%) for \textsc{SkillGraph} over training. NQ and HotpotQA are in-domain training datasets; the remaining datasets are held-out transfer evaluations.}
    \label{tab:search_curve}
    \centering
    \footnotesize
    \setlength{\tabcolsep}{3pt}
    \begin{tabular}{rrrrrrrrr}
    \toprule
    Step & NQ & TriviaQA & PopQA & HotpotQA & 2Wiki & MuSiQue & Bamboogle & Avg. \\
    \midrule
     30 & 39.2 & 58.9 & 40.6 & 39.7 & 35.4 & 16.3 & 69.0 & 42.4 \\
     60 & 44.9 & 61.8 & 44.1 & 42.4 & 39.1 & 17.6 & 69.8 & 45.3 \\
     90 & 46.0 & 62.1 & 45.6 & 42.8 & 40.0 & 18.1 & 68.1 & 46.4 \\
    120 & 46.2 & 62.5 & 45.1 & 43.6 & 41.9 & 18.4 & 70.6 & 46.9 \\
    150 & 46.4 & 63.6 & 45.6 & 44.3 & 43.3 & 19.4 & 70.6 & 47.9 \\
    180 & \textbf{48.3} & 63.4 & 47.0 & \textbf{44.8} & 41.9 & 19.4 & 72.2 & 48.1 \\
    200 & 48.0 & \textbf{63.8} & \textbf{48.5} & 44.7 & \textbf{43.4} & \textbf{19.5} & \textbf{72.6} & \textbf{48.9} \\
    \bottomrule
    \end{tabular}
\end{table}

\begin{table}[htbp]
  \caption{Licenses of datasets, environments, and models used in this work.}
  \label{tab:licenses}
  \centering
  \footnotesize
  \setlength{\tabcolsep}{3pt}
  \begin{tabular}{llll}
  \toprule
  Asset & Type & License & Reference \\
  \midrule
  \multicolumn{4}{l}{\textit{Environments}} \\
  ALFWorld & Environment & MIT & \citet{shridhar2020alfworld} \\
  WebShop & Environment & MIT & \citet{yao2022webshop} \\
  \midrule
  \multicolumn{4}{l}{\textit{Datasets --- Single-hop QA}} \\
  Natural Questions (NQ) & Dataset & Apache 2.0 & \citet{kwiatkowski2019nq} \\
  TriviaQA & Dataset & Apache 2.0 & \citet{joshi2017triviaqa} \\
  PopQA & Dataset & MIT & \citet{mallen2023popqa} \\
  \midrule
  \multicolumn{4}{l}{\textit{Datasets --- Multi-hop QA}} \\
  HotpotQA & Dataset & CC BY-SA 4.0 & \citet{yang2018hotpotqa} \\
  2WikiMultiHopQA & Dataset & Apache 2.0 & \citet{ho2020twowiki} \\
  MuSiQue & Dataset & CC BY 4.0 & \citet{trivedi2022musique} \\
  Bamboogle & Dataset & MIT & \citet{press2023bamboogle} \\
  \midrule
  \multicolumn{4}{l}{\textit{Models}} \\
  Qwen2.5-7B-Instruct & Model & Apache 2.0 & \citet{bai2023qwen} \\
  OpenAI o3 & API Service & Proprietary & \citet{openai2025o3} \\
  \bottomrule
  \end{tabular}
\end{table}

\section{Compute Resources}
\label{app:compute}

All training experiments are conducted on a single node equipped with 8$\times$ NVIDIA A100 80GB GPUs, 224 CPU cores, and 2048 GB of system memory. The total compute budget across all training runs amounts to approximately 280 GPU-hours.

\section{Broader Impact}
\label{app:broader_impact}

This work proposes a general framework for organizing and evolving reusable skills in LLM-based agents. We discuss potential broader impacts below.

\paragraph{Positive impacts.}
By enabling agents to accumulate structured knowledge from experience and reuse it across tasks, \textsc{SkillGraph} can improve the sample efficiency and reliability of autonomous agents in domains such as household assistance, web navigation, and information retrieval. The graph-structured skill memory also enhances interpretability: users can inspect which skills were retrieved, how they are related, and why certain decisions were made, facilitating human oversight of agent behavior. Furthermore, the progressive unlocking mechanism provides a built-in safety property---agents are restricted to well-mastered foundational skills before being exposed to more complex behaviors, reducing the risk of premature deployment of unreliable capabilities.

\paragraph{Potential risks and limitations.}
As with other LLM agent systems, \textsc{SkillGraph} inherits the biases and failure modes of the underlying language model. Skills distilled from trajectories may encode undesirable patterns if the training data contains biased behaviors. The teacher model used for graph evolution (e.g., skill insertion, merge, split) may introduce errors or hallucinated skills, which could propagate through the graph. We mitigate this through the deprecation mechanism that removes consistently failing skills, but additional safeguards (e.g., human-in-the-loop skill review) may be necessary for safety-critical applications. Our current evaluation focuses on simulated environments; deployment in real-world settings would require careful validation of skill quality and additional safety constraints.

\section{Asset Licenses}
\label{app:licenses}

Table~\ref{tab:licenses} summarizes the licenses of all datasets, environments, and models used in this work. All assets are publicly available and permit academic research use.

\section{Supplementary Experiments and Analysis}
\label{app:or_supplement}

This appendix examines the contributions of graph structure and maintenance, sensitivity to maintenance thresholds, and the computational cost of \textsc{SkillGraph}. We also discuss task-specific error patterns and the interaction between graph evolution and policy optimization. Unless otherwise stated, ALFWorld results are success rates (\%), WebShop results are task scores on a $0$--$100$ scale, and performance differences are absolute points.

\subsection{Controlled Graph Interventions}
\label{app:or_controls}

\paragraph{Intervention design.}
We examine four interventions that change skill organization or graph operations within the same base-model, cold-start SFT, and GRPO framework. The flat-library variant retains the skill content but removes explicit graph topology. The prerequisite-removal variant deletes prerequisite relations while retaining the other edge types. Retrieval without topology selects relevant skills without dependency expansion or topological ordering. The static-graph variant retains graph-based retrieval but disables graph evolution during training. These comparisons probe the roles of relational structure, dependency recovery, retrieval, and adaptation, respectively.

\begin{table}[htbp]
\caption{Graph interventions on ALFWorld success rate and WebShop task score. Drops are absolute differences from the full method in each column.}
\label{tab:or_controls}
\centering
\small
\setlength{\tabcolsep}{4pt}
\begin{tabular}{lrrrr}
\toprule
& \multicolumn{2}{c}{ALFWorld success} & \multicolumn{2}{c}{WebShop score}\\
Variant & Score & Drop & Score & Drop\\
\midrule
Full method & 93.1 & -- & 91.5 & --\\
Flat skill library & 89.9 & 3.2 & 85.2 & 6.3\\
Remove prerequisite edges & 87.5 & 5.6 & 82.5 & 9.0\\
Retrieval without topology & 59.4 & 33.7 & 83.3 & 8.2\\
Static graph & 78.2 & 14.9 & 70.3 & 21.2\\
\bottomrule
\end{tabular}
\end{table}

\paragraph{Results and interpretation.}
Table~\ref{tab:or_controls} shows that every intervention reduces performance in both environments. Replacing the graph with a flat skill library lowers ALFWorld success from $93.1$ to $89.9$ and WebShop score from $91.5$ to $85.2$, corresponding to drops of $3.2$ and $6.3$ points. Removing prerequisite edges incurs drops of $5.6$ and $9.0$ points, supporting the value of explicit dependencies alongside associative connections. On ALFWorld, retrieval without topology produces the largest decrease, $33.7$ points, consistent with the need to recover and order intermediate actions in household tasks. On WebShop, the static graph produces the largest decrease, $21.2$ points, suggesting that adapting skill content and relations is particularly useful when search and product-selection strategies must improve with the policy.

Together, these results highlight complementary roles for graph-aware retrieval and evolution. Dependency recovery and ordering provide structured guidance for ALFWorld's sequential tasks, while continued refinement of skills and relations supports WebShop's search and product-selection decisions.

\subsection{Joint Sensitivity to Maintenance Thresholds}
\label{app:or_sensitivity}

\paragraph{Settings.}
We use shared default thresholds across environments and examine two joint perturbations, denoted Loose and Strict in Table~\ref{tab:or_sensitivity}. Each setting changes the update, merge, deprecation, and level-unlock thresholds together while keeping the remaining settings fixed. A lower merge threshold allows more pairs of skills to qualify for merging, while a lower deprecation threshold retains low-success skills longer. Lowering the unlock threshold makes higher-level skills available earlier; raising it requires stronger evidence of competence at the current level. These joint settings examine how the balance between skill maintenance and progressive exposure affects performance.

\begin{table}[htbp]
\caption{Joint sensitivity to update, merge, deprecation, and unlock thresholds on ALFWorld success rate and WebShop task score.}
\label{tab:or_sensitivity}
\centering
\small
\setlength{\tabcolsep}{4pt}
\begin{tabular}{lrrrrrr}
\toprule
Setting & Update & Merge & Deprecate & Unlock & ALFWorld & WebShop\\
\midrule
Default & 0.40 & 0.85 & 0.15 & 0.60 & 93.1 & 91.5\\
Loose & 0.30 & 0.75 & 0.10 & 0.50 & 87.5 & 89.1\\
Strict & 0.60 & 0.95 & 0.20 & 0.70 & 89.8 & 85.7\\
\bottomrule
\end{tabular}
\end{table}

\paragraph{Results.}
The default configuration performs best among the three settings in both environments. Relative to the default, the loose setting reduces ALFWorld success by $5.6$ points and WebShop score by $2.4$ points; the strict setting reduces them by $3.3$ and $5.8$ points, respectively. Both perturbations remain above the skill-free GRPO reference in Table~\ref{tab:main}: the loose and strict settings exceed GRPO by $9.9$ and $12.2$ points on ALFWorld, and by $9.8$ and $6.4$ points on WebShop task score. Thus, the advantage over GRPO persists under both tested joint perturbations. ALFWorld is more sensitive to the loose setting, whereas WebShop is more sensitive to the strict setting, suggesting that the shared default provides an effective balance between skill refinement and progressive exposure across the two environments.

\subsection{Teacher Usage, Training Compute, and Efficiency}
\label{app:or_cost}
\paragraph{Teacher-model usage.}
The supplementary comparisons reuse the public SkillRL cold-start SFT data. During training, o3 supports failure analysis, skill insertion, and merge/split verification through calls triggered by graph-maintenance conditions. The policy and graph algorithms handle action generation, traversal, topological sorting, edge-weight updates, pruning, and final evaluation. Table~\ref{tab:or_teacher} reports average additional usage of $102.5$ calls on ALFWorld and $74.3$ calls on WebShop, costing US\$1.91 and US\$1.39, respectively. The combined incremental API cost for graph maintenance is US\$3.30.

\begin{table}[htbp]
\caption{Average additional o3 calls and API cost for graph maintenance on ALFWorld and WebShop.}
\label{tab:or_teacher}
\centering
\small
\begin{tabular}{lrr}
\toprule
Environment & Average o3 calls & Additional cost (USD)\\
\midrule
ALFWorld & 102.5 & 1.91\\
WebShop & 74.3 & 1.39\\
\midrule
Combined cost & -- & 3.30\\
\bottomrule
\end{tabular}
\end{table}

\paragraph{Training time and historical cost estimate.}
The supplementary comparison runs use four A100 GPUs each. Table~\ref{tab:or_compute} shows that \textsc{SkillGraph} reduces training time from $32.26$ to $31.59$ hours on ALFWorld and from $12.34$ to $11.39$ hours on WebShop, corresponding to reductions of $2.1\%$ and $7.7\%$. Across the two environments, training time decreases by $1.62$ hours, or $3.6\%$. To express compute and teacher usage in the same monetary units, we use the historical estimate of US\$21.9576 per hour for an eight-A100 instance, equivalent to US\$2.7447 per A100-hour. Estimated GPU cost is the training duration multiplied by four GPUs and this hourly rate.

\begin{table}[htbp]
\caption{Training wall-clock hours on four A100 GPUs and estimated GPU costs for the supplementary comparison runs.}
\label{tab:or_compute}
\centering
\small
\begin{tabular}{llrr}
\toprule
Environment & Method & Hours & Estimated GPU cost (USD)\\
\midrule
ALFWorld & SkillRL / flat skills & 32.26 & 354.18\\
ALFWorld & \textsc{SkillGraph} & 31.59 & 346.82\\
WebShop & SkillRL / flat skills & 12.34 & 135.48\\
WebShop & \textsc{SkillGraph} & 11.39 & 125.05\\
\midrule
Combined & SkillRL / flat skills & 44.60 & 489.65\\
Combined & \textsc{SkillGraph} & 42.98 & 471.87\\
\bottomrule
\end{tabular}
\end{table}

Relative to the estimated GPU costs of \textsc{SkillGraph}, additional teacher usage amounts to $0.55\%$ on ALFWorld, $1.11\%$ on WebShop, and $0.70\%$ across both environments. Local RL training therefore accounts for most of the estimated expense in these comparison runs, with graph-maintenance calls adding a small API cost.

\paragraph{Context and step efficiency.}
Graph-aware retrieval expands from task-relevant seeds and caps the resulting skill sequence, supplying a compact set of skills with their dependency context. Table~\ref{tab:or_efficiency} reports an average WebShop prompt length of $1{,}419$ tokens, compared with $1{,}781$ for flat skills, a reduction of $362$ tokens ($20.3\%$). The final step-token count decreases from $1.81$M to $1.37$M ($24.3\%$), while step time decreases from $302.9$ to $292.4$ seconds ($3.5\%$).

\begin{table}[htbp]
\caption{WebShop prompt length, step-token count, and step time for flat skills and \textsc{SkillGraph}.}
\label{tab:or_efficiency}
\centering
\small
\begin{tabular}{lrr}
\toprule
Measurement & SkillRL / flat skills & \textsc{SkillGraph}\\
\midrule
Average prompt length (tokens) & 1,781 & 1,419\\
Final step-token count & 1.81M & 1.37M\\
Step time (seconds) & 302.9 & 292.4\\
\bottomrule
\end{tabular}
\end{table}

The reductions in prompt length and token count are consistent with the shorter-context trend in Figure~\ref{fig:analysis_curves}. Together with the lower step time, they show that graph-aware retrieval improves context efficiency while reducing training-step runtime in the WebShop comparison.

\paragraph{Time measurements.}
Table~\ref{tab:or_time} summarizes the recorded time measurements for closed-source models and the two local RL methods. For the local RL comparison, the entries correspond to training hours: \textsc{SkillGraph} takes $42.98$ hours across ALFWorld and WebShop, compared with $44.60$ hours for SkillRL. The reduction in both environments complements the context-efficiency results in Table~\ref{tab:or_efficiency}.

\begin{table}[htbp]
\caption{Recorded time measurements on ALFWorld and WebShop. The SkillRL and \textsc{SkillGraph} entries report training hours.}
\label{tab:or_time}
\centering
\small
\begin{tabular}{lrrr}
\toprule
Method & ALFWorld & WebShop & Total\\
\midrule
GPT-4o & 13.88 & 2.24 & 16.12\\
Gemini-2.5-Pro & 28.82 & 22.45 & 51.27\\
SkillRL & 32.26 & 12.34 & 44.60\\
\textsc{SkillGraph} & 31.59 & 11.39 & 42.98\\
\bottomrule
\end{tabular}
\end{table}

\subsection{QA Clarifications and Qualitative Error Analysis}
\label{app:or_qualitative}
\paragraph{QA results and comparisons.}
Table~\ref{tab:qa} reports NQ and PopQA scores of $48.0$ and $48.5$ for \textsc{SkillGraph}, exceeding SkillRL by $2.1$ and $2.6$ points. ZeroSearch achieves the highest PopQA score of $51.5$, while \textsc{SkillGraph} achieves the highest arithmetic mean over the seven benchmarks at $48.6$, compared with $47.5$ for SkillRL. On MuSiQue, \textsc{SkillGraph} scores $19.5$, exceeding EvolveR's $15.6$ by $3.9$ points and falling below SkillRL's $20.2$ by $0.7$ points. On 2Wiki, \textsc{SkillGraph} achieves $43.4$, compared with EvolveR's $42.0$. These results show broad gains over SkillRL alongside variation across individual QA tasks.

The following qualitative examples illustrate task-specific failure mechanisms and the guidance that graph-organized skills can provide.

\paragraph{ALFWorld errors.}
ALFWorld requires the agent to track object locations, inventory, and completed subgoals across a sequence of constrained actions. State-tracking failures can lead an agent to continue searching after observing the target or to overlook an object inside a closed container. Execution failures include attempting to clean, heat, or cool an object before acquiring it, choosing the wrong device, or omitting final placement. A skill chain such as \emph{find $\to$ pick $\to$ transform $\to$ place} makes the dependency structure explicit and can help preserve the overall plan across local decisions. This procedural guidance complements the policy's interpretation of observations and tracking of environment state.

\paragraph{WebShop errors.}
WebShop requires satisfying several product constraints simultaneously. A product may match the requested category while violating a size, color, or price requirement, and an apparently suitable listing may still require selecting the correct option before purchase. A chain such as \emph{query refinement $\to$ filtering $\to$ option verification $\to$ final check} encourages the agent to retain all constraints through the interaction and verify the final selection. Graph evolution can refine these strategies as failures expose missing checks, which is consistent with the static-graph result in Table~\ref{tab:or_controls}.

\paragraph{MuSiQue difficulties.}
MuSiQue requires recovering implicit bridge entities and carrying them into subsequent queries to assemble a complete evidence chain. An incorrect intermediate entity can direct the next search toward irrelevant evidence, while incomplete retrieved evidence can leave the final answer unsupported even when the reasoning steps are ordered correctly. Under a limited search budget, successful execution depends on accurate entity selection, focused follow-up queries, and evidence integration. These demands provide a plausible explanation for the task's lower absolute scores and motivate skills that explicitly preserve intermediate entities and verify evidence coverage.

\subsection{Additional Methodological Details}
\label{app:or_scope}
\paragraph{Three relation types.}
The three relation types serve distinct operations in graph-aware retrieval. Prerequisite edges recover dependencies and determine execution order; enhancement edges connect general advice to task-specific procedures; and co-occurrence edges expand retrieval toward skills that have been useful together. Co-occurrence expresses associative relevance, while prerequisite and enhancement edges define the dependency levels described in Appendix~\ref{app:level_computation}. Together, these relations support dependency recovery, general-skill propagation, and associative expansion. The prerequisite-removal results in Table~\ref{tab:or_controls} further support the role of explicit ordering relations in both environments.

\paragraph{Meaning of co-evolution.}
Co-evolution refers to the feedback loop between policy learning and graph maintenance. GRPO updates the policy from rollouts conditioned on retrieved skills. Subsequent validation trajectories update skill-usage and success statistics, which guide changes to skill nodes, relation weights, and the active skill set. The revised graph then changes the guidance supplied to later policy updates. Teacher calls support selected skill-generation and verification operations within this loop. Policy optimization and discrete graph updates therefore interact through trajectory feedback and the skill guidance used in subsequent rollouts.

\paragraph{Experimental setting.}
The supplementary comparisons use Qwen2.5-7B-Instruct, keeping the base policy fixed when examining skill organization and maintenance. ALFWorld and WebShop evaluate the framework in sequential manipulation and multi-constraint navigation, respectively. The QA experiments train on NQ and HotpotQA and evaluate zero-shot on five additional datasets, assessing transfer of the search-augmented policy to unseen QA tasks.

\end{document}